\documentclass[conference]{IEEEtran}
\IEEEoverridecommandlockouts
\usepackage{cite}
\usepackage{amsmath,amssymb,amsfonts}
\usepackage{graphicx}
\usepackage{textcomp}
\usepackage{xcolor}
\usepackage[nolist]{acronym}
\def\BibTeX{{\rm B\kern-.05em{\sc i\kern-.025em b}\kern-.08em
    T\kern-.1667em\lower.7ex\hbox{E}\kern-.125emX}}
\usepackage{algorithm}
\usepackage{algorithmicx}
\usepackage{algpseudocode}
\usepackage{booktabs}
\usepackage{fontawesome}
\usepackage{multirow}
\usepackage{url}

\begin{document}
\acrodef{3GPP}{3rd Generation Partnership Project}
\acrodef{5G}{5th Generation Mobile Network}
\acrodef{6G}{6th Generation Mobile Network}
\acrodef{AI}{Artificial Intelligence}
\acrodef{AI4Net}{\ac{AI} for Networking}
\acrodef{MDP}{Markov Decision Process}
\acrodef{AIDER}{Aerial Image Dataset for Emergency Response}
\acrodef{AMF}{Access and Mobility Management Function}
\acrodef{AIaaS}{Artificial Intelligence-as-a-Service}
\acrodef{AC}{Actor-Critic}
\acrodef{AIDER}{Aerial Image Database for Emergency Response applications}
\acrodef{B5G}{Beyond Fifth Generation}
\acrodef{BPF}{Berkeley Packet Filter}
\acrodef{CBR}{Constant Bit Rate}
\acrodef{CSV}{Comma-Separated Values}
\acrodef{CPU}{Central Processing Unit}
\acrodef{CNN}{Convolutional Neural Network}
\acrodef{CNNs}{Convolutional Neural Networks}
\acrodef{C-V2X}{Cellular Vehicle to-Everything}
\acrodef{DoS}{Denial of Service}
\acrodef{DDQL}{Double Deep
 Q-learning}
\acrodef{DDoS}{Distributed Denial of Service}
\acrodef{DDPG}{Deep Deterministic Policy Gradient}
\acrodef{DNN}{Deep Neural Network}
\acrodef{DRL}{Deep Reinforcement Learning}
\acrodef{DQN}{Deep Q-Network}
\acrodef{DT}{Decision Tree}
\acrodef{DDQN}{Double Deep Q-Network}

\acrodef{ETSI}{European Telecommunications Standards Institute}
\acrodef{eNWDAF}{Evolved Network Data Analytics Function}
\acrodef{eBPF}{Extended Berkeley Packet Filter}
\acrodef{ECDF}{Empirical Cumulative Distribution Function}
\acrodef{ECDFs}{Empirical Cumulative Distribution Functions}
\acrodef{FIBRE}{Future Internet Brazilian Environment for Experimentation}
\acrodef{FL}{Federated Learning}
\acrodef{FEL}{Federated Ensemble Learning}
\acrodef{GNN}{Graph Neural Networks}
\acrodef{GPU}{Graphics Processing Unit}
\acrodef{GTP}{GPRS Tunnelling Protocol}
\acrodef{GTP-U}{GPRS Tunnelling Protocol User Plane}
\acrodef{GA}{Genetic Algorithm}
\acrodef{HTM}{Hierarchical Temporal Memory}

\acrodef{IAM}{Identity And Access Management}
\acrodef{ICMP}{Internet Control Message Protocol}
\acrodef{IID}{Independent and Identically Distributed}
\acrodef{IoE}{Internet of Everything}
\acrodef{IoT}{Internet of Things}
\acrodef{ITU}{International Telecommunication Union}
\acrodef{IQR}{Interquartile Range}
\acrodef{I/O}{Input/Output}
\acrodef{IP}{Internet Protocol}
\acrodef{KNN}{K-Nearest Neighbors}
\acrodef{KPI}{Key Performance Indicator}
\acrodef{KPIs}{Key Performance Indicators}
\acrodef{LSTM}{Long Short-Term Memory}
\acrodef{LOWESS}{Locally Weighted Scatterplot Smoothing}
\acrodef{LR}{Learning Rate}
\acrodef{MAE}{Mean Absolute Error}
\acrodef{MAD}{Median Absolute Deviation}
\acrodef{ML}{Machine Learning}
\acrodef{MLaaS}{Machine Learning as a Service}
\acrodef{MOS}{Mean Opinion Score}
\acrodef{MAPE}{Mean Absolute Percentage Error}
\acrodef{MSE}{Mean Squared Error}
\acrodef{MEC}{Multi-access Edge Computing}
\acrodef{mMTC}{Massive Machine Type Communications}
\acrodef{MFA}{Multi-factor Authentication}
\acrodef{MLP}{Multi-Layer Perceptron}
\acrodef{MADRL}{Multi-Agent Deep Reinforcement Learning}
\acrodef{MAB}{Multi-Armed Bandit}
\acrodef{MILP}{Mixed Integer Linear Programming}
\acrodef{MQTT}{Message Queuing Telemetry Transport}
\acrodef{NWDAF}{Network Data Analytics Function}
\acrodef{Net4AI}{Networking for \ac{AI}}
\acrodef{NS}{Network Slicing}
\acrodef{NFV}{Network Function Virtualization}
\acrodef{OSM}{Open Source MANO}
\acrodef{PCA}{Principal Component Analysis}
\acrodef{PC-FedAvg}{Personalized Conditional Federated Averaging}
\acrodef{PoC}{Proof of Concept}
\acrodef{PPO}{Proximal Policy Optimization}
\acrodef{POMDP}{Partially Observable Markov decision process}
\acrodef{PCAP}{Packet Capture}
\acrodef{PSO}{Particle Swarm Optimization}
\acrodef{QoE}{Quality of experience}
\acrodef{QoS}{Quality of Service}
\acrodef{QFI}{QoS Flow Identifier}
\acrodef{QFIs}{QoS Flow Identifiers}
\acrodef{RAM}{Random Access Memory}
\acrodef{RF}{Random Forest}
\acrodef{RL}{Reinforcement Learning}
\acrodef{RMSE}{Root Mean Square Error}
\acrodef{RNN}{Recurrent Neural Network}
\acrodef{RTT}{Round-Trip Time}
\acrodef{RAN}{Radio Access Network}
\acrodef{RTP}{Real-time Transport Protocol}
\acrodef{SDN}{Software-Defined Networking}
\acrodef{SFI2}{Slicing Future Internet Infrastructures}
\acrodef{SLA}{Service-Level Agreement}
\acrodef{SON}{Self-Organizing Network}
\acrodef{SMF}{Session Management Function}
\acrodef{S-NSSAI}{Single Network Slice Selection Assistance Information}
\acrodef{SVM}{Support Vector Machine}
\acrodef{SOPS}{Service-Aware Optimal
 Path Selection}
 \acrodef{SAFE}{Scalable Asynchronous Federated Ensembling}
\acrodef{TQFL}{Trust Deep Q-learning Federated Learning}
\acrodef{TEID}{Tunnel Endpoint Identifier}
\acrodef{TEIDs}{Tunnel Endpoint Identifiers}
\acrodef{TPE}{Tree-Structured Parzen Estimator}
\acrodef{UE}{User Equipment}
\acrodef{UEs}{User Equipments}
\acrodef{UPF}{User Plane Function}
\acrodef{UPFs}{User Plane Functions}
\acrodef{PDU}{Packet Data Unit}
\acrodef{URLLC}{Ultra-Reliable and Low Latency Communications}
\acrodef{UAV}{Unmanned Aerial Vehicle}
\acrodef{UAVs}{Unmanned Aerial Vehicles}
\acrodef{UDP}{User Datagram Protocol}
\acrodef{VoD}{Video on Demand}
\acrodef{VR}{Virtual Reality}
\acrodef{AR}{Augmented Reality}
\acrodef{V2V}{Vehicle-to-Vechile}
\acrodef{V2X}{Vehicle-to-Everything}
\acrodef{VNF}{Virtual Network Function}
\acrodef{VNFs}{Virtual Network Functions}

\acrodef{XDP}{eXpress Data Path}
\title{Asynchronous Probability Ensembling for Federated Disaster Detection}

\author{\IEEEauthorblockN{Emanuel Teixeira Martins\textsuperscript{1}, Rodrigo Moreira\textsuperscript{1}, Larissa Ferreira {Rodrigues Moreira}\textsuperscript{1},\\ Rodolfo S. Villaça\textsuperscript{2}, Augusto Neto\textsuperscript{3},
Flávio de Oliveira Silva\textsuperscript{4}}
\IEEEauthorblockA{
\textsuperscript{1}Federal University of Viçosa (UFV), Rio Paranaíba-MG, Brazil\\
\textsuperscript{2}Federal University of Espirito Santo (UFES), Vitória-ES, Brazil\\
\textsuperscript{3}Federal University of Rio Grande do Norte (UFRN), Natal-RN, Brazil\\
\textsuperscript{4}University of Minho (UMinho), Braga, Portugal\\
Emails: \{emanuel.martins, rodrigo, larissa.f.rodrigues\}@ufv.br,\\ rodolfo.villaca@ufes.br, augusto@dimap.ufrn.br, flavio@di.uminho.pt}
}

\maketitle

\begin{abstract}
Quick and accurate emergency handling in Disaster Decision Support Systems (DDSS) is often hampered by network latency and suboptimal application accuracy. While Federated Learning (FL) addresses some of these issues, it is constrained by high communication costs and rigid synchronization requirements across heterogeneous convolutional neural network (CNN) architectures. To overcome these challenges, this paper proposes a decentralized ensembling framework based on asynchronous probability aggregation and feedback distillation. By shifting the exchange unit from model weights to class-probability vectors, our method maintains data privacy, reduces communication requirements by orders of magnitude, and improves overall accuracy. This approach enables diverse CNN designs to collaborate asynchronously, enhancing disaster image identification performance even in resource-constrained settings. Experimental tests demonstrate that the proposed method outperforms traditional individual backbones and standard federated approaches, establishing a scalable and resource-aware solution for real-time disaster response.
\end{abstract}


\section{Introduction}\label{sec:introduction}

Timely and reliable identification of disasters from field images is a critical requirement in emergency response, particularly when sensing is performed by heterogeneous edge devices, such as \ac{UAVs} and smartphones, under intermittent connectivity~\cite{Lin2025, Moreira2025}. In these environments, network resources are constrained by damaged infrastructure, whereas devices vary significantly in computational and storage capacity. \ac{FL} offers a natural way to leverage decentralized data without centralizing sensitive content. However, deployments face persistent limitations, including non-\ac{IID} data, class imbalance, detection of selfish learners, and costly parameter exchange~\cite{Ribeiro2026, Imandi2025, Gecer2024}. These issues are particularly damaging in disaster response scenarios, where delayed or biased predictions can significantly affect decision-making.

Although \ac{FL} pioneered a vital privacy-preserving paradigm~\cite{McMahan2017}, its reliance on synchronized weight sharing is ill-suited for disaster identification. This core mechanism is impractical for fragile networks and vulnerable to biases from unreliable clients~\cite{Gecer2024}. Traditionally, layering ensemble methods onto this framework worsens the communication bottleneck by requiring the sharing of entire models ~\cite{Mabrouk2023, Casado2023}. This suggests that the core \ac{FL} idea of client collaboration is highly beneficial, not only for the costly exchange of model weights but also as an inspiration for lightweight ensemble methods. This approach avoids the high synchronization and communication costs that render conventional techniques ineffective for time-critical disaster responses.

To address this challenge, we propose a probability-level training scheme that aggregates softmax vectors instead of gradients or weights. Clients publish class-probability outputs to a lightweight \ac{MQTT} broker, and the server consumes them asynchronously to learn either a stacking meta-classifier or optimized combination weights. Operating at the prediction level, our method tolerates heterogeneous \ac{CNN} backbones, reduces payload size, and remains robust to client drop-offs by avoiding global synchronization. It also includes a knowledge distillation feedback loop that lets local models refine predictions from the aggregated ensemble distribution. Our novelty lies in the design of training and coordination, not in a new backbone or classifier, enabling asynchronous collaboration in the probability space without sharing raw data or model parameters.

Although \ac{FL} extensions address non-\ac{IID} data, class imbalance, and heterogeneity, many deployments still depend on synchronized rounds and repeated parameter exchange, which is costly and brittle under intermittent connectivity. We do not replace these methods; instead, we complement them by exchanging class probability vectors rather than parameters, enabling asynchronous participation and architectural heterogeneity with much lower payloads.

The main contributions of this study are as follows: (i) we identify the key limitations of \ac{FL} in disaster response, including synchronization overhead, vulnerability to poisoned or biased updates, and high communication cost; (ii) we propose an asynchronous probability-level aggregation framework that preserves privacy, reduces network load, and leverages architectural heterogeneity; (iii) we integrate ensemble strategies, stacking, \ac{GA}, \ac{PSO}, and a probability-level distillation loop, so local models benefit from global consensus without weight exchange; and (iv) we evaluate the \ac{AIDER} dataset with multiple \ac{CNN} backbones, showing accuracy comparable to or exceeding centralized and federated approaches while drastically reducing communication overhead.

The remainder of this paper is organized as follows. Section~\ref{sec:related_work} reviews prior work on ensemble learning in distributed settings. Section~\ref{sec:proposed_method} describes our method and distillation approach, Section~\ref{sec:evaluation_setup} details the testbed, Section~\ref{sec:results_and_discussion} reports results, and Section~\ref{sec:concluding_remarks} concludes the paper. 

\section{Related Work}\label{sec:related_work}

This section reviews related work on federated ensembling and communication-efficient learning, situating our method as an asynchronous, probability-level alternative.

Mabrouk et al.~\cite{Mabrouk2023} proposed an ensemble \ac{FL} for pneumonia detection, where hospitals trained multiple \ac{CNNs} and shared local ensembles for global aggregation. Unlike this parameter-level exchange, our approach aggregates softmax vectors asynchronously, reducing communication and enabling heterogeneous participation.

Gao et al.~\cite{Gao2024} investigated cognitive data fusing for \ac{IoT} prognostics, combining \ac{FEL} with pruning and adaptive selection to improve resilience. Our approach instead performs direct probability-level aggregation, reducing synchronization costs while supporting diverse \ac{CNN} backbones for disaster response.

Zhao et al.~\cite{Zhao2024} proposed a byzantine-robust ensemble incentive mechanism for \ac{FL}, building multiple global models and aggregating them with a distance-based rule while rewarding honest clients. While their focus is on adversarial robustness at the model level, our method operates in the probability space via asynchronous softmax aggregation to enable efficient disaster identification.

Sievers et al.~\cite{Sievers2024} applied a mixture-of-experts architecture in \ac{FL} for smart grid energy forecasting, improving accuracy and reducing training time under non-\ac{IID} conditions. Akram et al.~\cite{Akram2025} introduced \ac{PC-FedAvg}, a conditional \ac{FL} scheme for post-disaster structural health monitoring. By selectively aggregating client models with low validation loss, their method improves robustness for non-\ac{IID} data. Our method, however, broadens scalability by shifting aggregation from weights to probability vectors, making it lightweight and tolerant to architectural heterogeneity.

Xu et al.~\cite{Xu2025} presented an adaptive sampling-based \ac{FEL} for aircraft remaining functional life prediction. By combining multiple deep models and prioritizing clients with lower data quality, the robustness under heterogeneous conditions was enhanced. Our approach follows a different path, employing asynchronous probability-level fusion to achieve efficiency in bandwidth-constrained disaster identification.

Hajla et al.~\cite{Hajla2025} proposed a hybrid \ac{FEL} for intrusion detection in \ac{IoT} networks, combining traditional classifiers via majority voting to improve accuracy and resilience. Although effective, reliance on conventional models limits representational capacity, whereas our method exploits deep \ac{CNN} backbones and asynchronous probability aggregation to achieve scalable, communication-efficient disaster identification.

Sheth et al.~\cite{Sheth31122024} propose an ensemble for natural disaster image classification that fuses an InceptionV3-based model with a custom CNN, trained on a Kaggle dataset with cyclone, wildfire, flood, and earthquake classes. It reports that the ensemble outperforms the individual models, achieving about 92.79 percent accuracy for this multi-class setting.

Wong et al.~\cite{Wong2022} propose an optimized multi-task learning model that jointly performs disaster classification and victim detection by attaching a pruned head to a backbone network, and by selecting the branching location and the pruned head depth. It evaluates this design in federated learning environments using FedAvg and production tools such as OpenFL and OpenVINO, showing reduced memory requirements and approximately 1-2\% higher classification accuracy while maintaining comparable detection performance.

Table \ref{tab:comparison} summarizes the related work with these columns: ``Field'' indicates the problem domain; \ac{FEL} (prob.) states whether it exchanges prediction outputs, such as probabilities or logits; Client Selection indicates whether a strategy actively chooses clients. This is a design choice of our approach: by dispensing with client selection, our strategy gains robustness and fairness, naturally absorbing intermittent participation while avoiding coordination and subset-management overhead; ``Async. Training'' highlights whether the method supports asynchronous participation without global synchronization; ``Heterogeneous Models'' shows the ability to combine different \ac{CNN} backbones; and \ac{AI} Models lists which models are used.

Beyond federated ensembling, our approach aligns with decentralized classification, where agents send belief vectors to a fusion center using a maximum a posteriori rule. Since disaster imagery from heterogeneous sensors requires deep, calibrated representations, our method acts as a practical probability‑level fusion and feedback mechanism for deep visual pipelines under intermittent connectivity.

\begin{table}[!t]
\centering
\caption{Comparison with related work. The symbols denote full support ({\tiny \faCircle}), partial or adaptive support ({\tiny \faAdjust}), and no support ({\tiny \faCircleO}).}
\label{tab:comparison}
\resizebox{\columnwidth}{!}{%
\begin{tabular}{@{}ccccccc@{}}
\toprule
\textbf{Work}      & \textbf{Field}                                                  & \textbf{\begin{tabular}[c]{@{}c@{}}FEL \\ (prob.)\end{tabular}} & \textbf{\begin{tabular}[c]{@{}c@{}}Client \\ Selection\end{tabular}} & \textbf{\begin{tabular}[c]{@{}c@{}}Async. \\ Training\end{tabular}} & \textbf{\begin{tabular}[c]{@{}c@{}}Heterogeneous \\ Models\end{tabular}} & \textbf{\begin{tabular}[c]{@{}c@{}}AI \\ Models\end{tabular}}     \\ \midrule
\cite{Mabrouk2023} & Healthcare                                                            & \faCircleO                                                      & \faCircleO                                                           & \faCircleO                                                          & \faAdjust                                                               & Multiple CNNs                                                     \\ \midrule
\cite{Gao2024}     & Healthcare                                                            & \faCircle                                                       & \faAdjust                                                            & \faCircleO                                                          & \faCircleO                                                               & ResNet-20                                                         \\ \midrule
\cite{Zhao2024}    & Adversarial FL                                                        & \faCircleO                                                      & \faCircle                                                            & \faCircleO                                                          & \faCircleO                                                               & Multiple CNNs                                                     \\ \midrule
\cite{Sievers2024} & \begin{tabular}[c]{@{}c@{}}Smart \\ Grids\end{tabular}                & \faCircleO                                                      & \faCircle                                                            & \faCircleO                                                          & \faCircleO                                                               & \begin{tabular}[c]{@{}c@{}}LSTM, CNN, \\ Transformer\end{tabular} \\ \midrule
\cite{Akram2025}   & \begin{tabular}[c]{@{}c@{}}Infrastructure \\ Monitoring\end{tabular}  & \faCircleO                                                      & \faCircle                                                            & \faCircleO                                                          & \faCircleO                                                               & EfficientNet                                                      \\ \midrule
\cite{Xu2025}      & Aviation                                                              & \faCircleO                                                      & \faCircle                                                            & \faCircleO                                                          & \faCircleO                                                               & LSTM, CNN                                                         \\ \midrule
\cite{Hajla2025}   & IoT Security                                                          & \faAdjust                                                       & \faCircleO                                                           & \faCircleO                                                          & \faCircleO                                                               & \begin{tabular}[c]{@{}c@{}}MLP, RF, \\ XGBoost\end{tabular}       \\ \midrule
\cite{Sheth31122024}             & \begin{tabular}[c]{@{}c@{}}Disaster \\ Response\end{tabular}          & \faCircleO                                                      & \faCircleO                                                           & \faCircleO                                                          & \faCircle                                                                & \begin{tabular}[c]{@{}c@{}}InceptionV3,\\ CNN\end{tabular}         \\ \midrule
\cite{Wong2022}            & \begin{tabular}[c]{@{}c@{}}Disaster \\ Response\end{tabular}          & \faCircleO                                                      & \faAdjust                                                            & \faCircleO                                                          & \faCircleO                                                               & \begin{tabular}[c]{@{}c@{}}YOLOv3,\\ MobileNetV2\end{tabular}      \\ \midrule
\textbf{Ours}      & \textbf{\begin{tabular}[c]{@{}c@{}}Disaster \\ Response\end{tabular}} & \textbf{\faCircle}                                              & \textbf{\faCircleO}                                                  & \textbf{\faCircle}                                                  & \textbf{\faCircle}                                                       & \textbf{Multiple CNNs}                                            \\ \bottomrule
\end{tabular}
}
\end{table}


\textbf{Contribution Positioning}. Several communication-efficient and heterogeneity-aware \ac{FL} methods, such as FedProx~\cite{Li2020} and SCAFFOLD~\cite{Karimireddy2020}, remain important baselines when parameter exchange is feasible. Related federated distillation and logit-sharing families, such as FedMD and FedDF, are also relevant points of reference for probability-level collaboration. \textcolor{black}{Our contribution is positioned as a lightweight asynchronous alternative, not as a comprehensive empirical replacement for all these families under matched budgets.}

\section{Proposed Method}\label{sec:proposed_method}

In this section, we present the rationale of our approach, an asynchronous method for probability aggregation in distributed training scenarios.

\subsection{Asynchronous Architecture}\label{subsec:overal_architecture}

\begin{figure}[ht]
\centering
  \includegraphics[width=1\columnwidth]{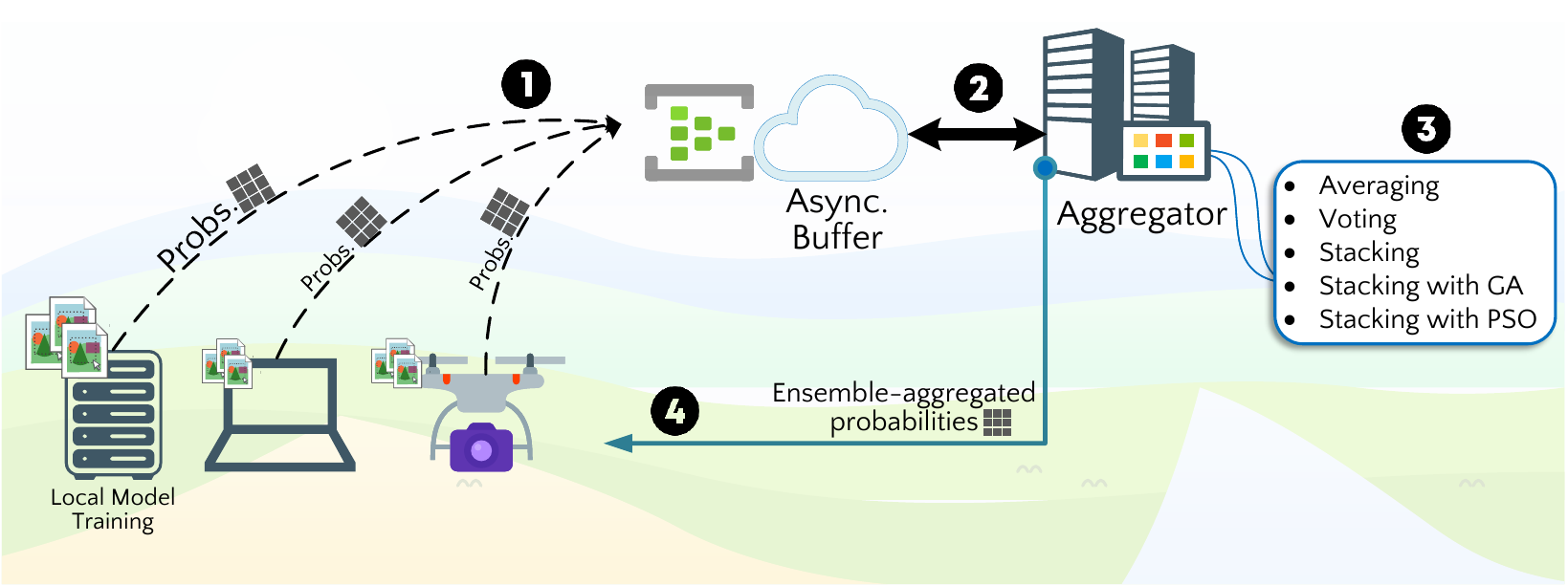}
  \caption{General overview of our method.}
  \label{fig:proposed_method}
\end{figure}

Our method is a probability-level, \ac{FL}-inspired training scheme designed to enhance communication efficiency, scalability, and robustness. Rather than exchanging gradients or model weights standards in \ac{FL}~\cite{mcmahan2017communication}, clients publish only their softmax probability vectors via an \ac{MQTT} broker. These lightweight vectors are aggregated asynchronously at the server. 
Clients may join at any time; if a client fails to send outputs in a round, training is not interrupted, and the server aggregates the available information. 
Optionally, the server broadcasts the stacked (aggregated) predictor to clients, enabling them to benefit from global knowledge without sharing raw data or parameters. Our method is depicted in Fig. ~\ref{fig:proposed_method}.


In the initial phase (1) of our methodology, clients train their models locally using their own architectures and publish the probabilities in queues. In phase two (2), the aggregator collects the probability vectors upon reaching a minimum number of contributions. During phase three (3), our approach employs various stacking methods to achieve higher accuracy on the test set. Finally, in phase four (4), the server returns the probability vector to the clients, enabling them to train their local models by targeting its loss and backpropagation to the consensus probabilities received from the server.

In phase three (3), we applied our ensemble learning method in three variants. In the first step, known as stacking, the probability vectors from each model are concatenated into a feature vector and used as input to a logistic regression meta-classifier. In the second and third variants, we explored an optimization-based approach that improves accuracy over baselines while reducing communication and enabling optimal combination weights to the client models, but differ in their search strategies: \ac{GA} relies on evolutionary operators such as crossover and mutation. In contrast, \ac{PSO} simulates the cooperative behavior of particles adjusting their positions according to inertia, self-knowledge, and swarm knowledge. In all cases, the fitness function used to guide the search was the classification accuracy.


\subsection{Knowledge Distillation Feedback Loop}\label{subsec:kd_feedback}

Our approach operates in accordance with Algorithm~\ref{alg:feedback}, where the server returns an updated set of model weights to the clients. Our method (Fig.~\ref{fig:proposed_method}) enables a complementary mechanism at the level of probabilities. After aggregating the clients' probability vectors through ensemble strategies (averaging, stacking, \ac{GA}, \ac{PSO}), the server produces an ensemble probability distribution $P_{\text{ensemble}}$. This distribution can be interpreted as a ``soft label'' or distilled knowledge~\cite{Hinton2015}, representing a consensus that is typically more accurate and robust than the output of any single client.

To make probability aggregation and distillation well defined, aggregated probability vectors must correspond to the same inputs. Our approach therefore operates over a shared reference set $X_{\mathrm{ref}}$, for example a public calibration split, a small set of field samples replicated across nodes, or a list of sample identifiers agreed upon by the server, such as hashes or timestamps. In each round, the server announces $X_{\mathrm{ref}}$ or its identifiers, each client computes probabilities only for these inputs, and the server aggregates probabilities per sample, producing $P_{\mathrm{ens}}(x)$ for each $x \in X_{\mathrm{ref}}$. The server then broadcasts these per sample distributions back to clients, enabling a consistent distillation objective.

\begin{algorithm}[htpb]
\scriptsize
\caption{Knowledge Distillation with a Shared Reference Set.}
\label{alg:feedback}
\begin{algorithmic}[1]
\Require Clients $\{C_1, \ldots, C_K\}$, local models $f_i$, local datasets $D_i$, reference set $X_{\mathrm{ref}}$
\Repeat
    \State Server broadcasts identifiers of $X_{\mathrm{ref}}$
    \For{each client $C_i$ \textbf{in parallel}}
        \State Train $f_i$ on local data $D_i$
        \State Compute probabilities for the reference set,
        \[
            P^{(i)}(x) = f_i(x), \quad \forall x \in X_{\mathrm{ref}}
        \]
        \State Publish $\{P^{(i)}(x)\}_{x \in X_{\mathrm{ref}}}$ to the server
    \EndFor
    \State Server aggregates per sample into $P_{\mathrm{ens}}(x)$ for all $x \in X_{\mathrm{ref}}$
    \State Server broadcasts $\{P_{\mathrm{ens}}(x)\}_{x \in X_{\mathrm{ref}}}$ back to clients
    \For{each client $C_i$ \textbf{in parallel}}
        \State Update $f_i$ by minimizing the distillation loss over $X_{\mathrm{ref}}$,
        \[
            \mathcal{L}_{KD} = \sum_{x \in X_{\mathrm{ref}}}\sum_{c=1}^{C}
            P_{\mathrm{ens}}(c \mid x)\,\log\frac{P_{\mathrm{ens}}(c \mid x)}{P^{(i)}(c \mid x)}
        \]
    \EndFor
\Until{convergence or maximum rounds reached}
\end{algorithmic}
\end{algorithm}

In the remainder of this section, $K$ denotes the number of clients and $C$ denotes the number of classes, avoiding overloaded symbols.

Each client may then refine its local model by aligning its predictions $P_{\text{local}}$ with $P_{\text{ensemble}}$ using a distillation objective~\cite{Hinton2015}. Concretely, for a given input sample $x$, the local training objective becomes to minimize the Kullback–Leibler divergence between the two distributions (Eq.~\ref{eq:kd_loss}):
\begin{equation}
\label{eq:kd_loss}
\mathcal{L}_{KD} = 
\sum_{x \in X_{\mathrm{ref}}}
\sum_{c=1}^{C} 
P_{\mathrm{ens}}(c \mid x)\,
\log\frac{P_{\mathrm{ens}}(c \mid x)}{P_{\mathrm{local}}(c \mid x)}.
\end{equation}

where $C$ is the number of classes. This objective encourages the local model to internalize not only the final class decision but also the confidence structure across classes. Below we present the Algorithm~\ref{alg:feedback} that described our proposed closed-loop. 
In this way, our method not only aggregates knowledge but also redistributes it, acting as a teacher in a knowledge distillation process. 

\subsection{Asynchronous Stacking Methods}\label{subsec:stacking_methods}

For the stacking procedure, we used the probability distributions produced by each baseline model (in phase 1) as input for a meta-classifier. Concretely, given three models $m_1, m_2, m_3$ with softmax outputs $p^{(m_i)} \in \mathbb{R}^C$, where $C$ is the number of classes, we built the feature vector.
\[
x = \big[ p^{(m_1)}, p^{(m_2)}, p^{(m_3)} \big] \in \mathbb{R}^{3C}.
\]
This vector was then used to train a logistic regression classifier $f_\theta(x)$, which in Scikit-learn is optimized with cross-entropy loss and L2 regularization by default. The training was conducted using probabilities extracted from the training partition of the \ac{AIDER} dataset, and the evaluation was performed on the validation partition. In this way, the ensemble learned to weight the models without accessing the test folder directly. Each experiment tested all possible combinations of three models among five candidates (EfficientNet, MobileNetV2, ResNet, MobileNetV3, and SqueezeNet), yielding 10 ensembles. The logistic regression was trained with up to 1000 iterations and used parallel fitting to accelerate convergence. As a result, we collected accuracy, confusion matrices (absolute and normalized), and detailed classification reports.  

In the case of the \ac{GA}, our goal was to find the best set of weights to combine the ensemble's models. Each solution in the population was a vector of weights $w = (w_1, w_2, \ldots, w_M)$, normalized so that the sum of the weights was equal to one. The population was initialized with random weights drawn from a Dirichlet distribution to increase diversity. The fitness of each individual was measured using the classification accuracy on a validation set. We retained the top five individuals in each generation (elitism) and generated offspring via one-point crossover. Mutation was applied with a probability of $p_m = 0.3$ by adding Gaussian noise to one of the weights. Every ten generations, new random individuals were injected to prevent the population from becoming stuck in local optima. The algorithm was run for 100 generations with a population size of 40 individuals. Table~\ref{tab:ga_params} shows the parameters used in our experiments.  

\begin{table}[!t]
\centering
\caption{Genetic Algorithm Parameters.}
\label{tab:ga_params}
\begin{tabular}{ll}
\hline
\textbf{Parameter} & \textbf{Value} \\ \hline
Population size     & 40 \\
Generations         & 100 \\
Selection strategy  & Elitism (top 5 preserved) \\
Crossover           & One-point \\
Mutation rate ($p_m$)  & 0.3 (Gaussian noise) \\
Diversity injection & Every 10 generations \\
Fitness function    & Validation accuracy \\ \hline
\end{tabular}
\end{table}

For particle swarm optimization (\ac{PSO}), the objective was similar: to learn the optimal weights for combining the models. In PSO, each particle is also a weight vector $w = (w_1, w_2, \ldots, w_M)$ that is updated based on its own experience and the best solution found by the swarm. The velocity update rule is
\[
v_i(t+1) = \omega v_i(t) + c_1 r_1 \big(p_i - x_i(t)\big) + c_2 r_2 \big(g - x_i(t)\big),
\]
where $\omega$ is the inertia coefficient, $c_1$ and $c_2$ are the cognitive and social factors, $r_1,r_2 \sim U(0,1)$ are random numbers, $p_i$ is the best position found by particle $i$, and $g$ is the global best. In practice, this means that particles try to balance exploring new solutions with following the best-known ones. We used a swarm of 20 particles for 100 iterations, with $\omega = 0.7$, $c_1 = 1.5$, and $c_2 = 1.5$. The fitness function, as in GA, was the classification accuracy on the validation set. Table~\ref{tab:pso_params} shows the PSO configuration.  

\begin{table}[!t]
\centering
\caption{Particle Swarm Optimization Parameters.}
\label{tab:pso_params}
\begin{tabular}{lc}
\hline
\textbf{Parameter} & \textbf{Value} \\ \hline
Swarm size         & 20 \\
Iterations         & 100 \\
Inertia coefficient ($\omega$) & 0.7 \\
Cognitive factor ($c_1$) & 1.5 \\
Social factor ($c_2$) & 1.5 \\
Fitness function   & Validation accuracy \\ \hline
\end{tabular}
\end{table}

\section{Evaluation Setup}\label{sec:evaluation_setup}

To assess our asynchronous ensemble learning method, we conducted various experiments using \ac{CNNs} to extract features from the \ac{AIDER} dataset~\cite{Kyrkou2020}. It contains images across five classes: fire/smoke, floods, collapsed structures, traffic accidents, and normal scenarios. The dataset is intentionally imbalanced to reflect real-world conditions, with images of normal scenes significantly outnumbering disaster events. Our idea was to address one of the main challenges of \ac{FL}: how to aggregate models with different tensor structures. Instead of forcing weight aggregation, we worked at the probability vector level, which allowed us to apply other techniques to improve accuracy in more complex scenarios. As a starting point, we trained baseline models individually using a \ac{TPE} hyperparameter search. 

Our experiments were conducted on a workstation with an NVIDIA RTX 4000 Ada GPU. The framework was built in Python using PyTorch for the models and Scikit-learn for the meta-classifier. For asynchronous communication, we deployed an Eclipse Mosquitto \ac{MQTT} broker in Docker to ensure reproducibility. We first fine-tuned several \ac{CNN} models using pre-trained weights and tuned hyperparameters. During evaluation, each model published its softmax probabilities via \ac{MQTT} to a central server, where we tested three aggregation methods: stacking, a \ac{GA}, and \ac{PSO}. We trained five \ac{CNNs} (EfficientNet, MobileNetV2, ResNet, MobileNetV3, and SqueezeNet) with the best batch size, learning rate, and optimizer (Adam/SGD) using Hyperopt with \ac{TPE}. These models were trained separately and used as a standard for comparison. Our implementation and scripts to reproduce the experiments are publicly available\footnote{\url{https://github.com/romoreira/NetAI-AppEnsembeLearning/tree/mqtt-updated}}.

\section{Results and Discussion}\label{sec:results_and_discussion}

To validate our approach, we initially trained five \ac{CNN} backbones in a centralized manner to establish baselines. Subsequently, we trained the same models and configurations using the \ac{FL} approach to serve as a second benchmark. For all our training, we used the \ac{AIDER} dataset \textcolor{black}{partitioned into 80/20} and employed a \ac{TPE} search to select hyperparameters for each model, as detailed in Table~\ref{tab:hyperparameters}. Ultimately, we trained the same models using our \ac{FEL} approach \textcolor{black}{using two clients over three communication rounds}. 

\begin{table}[!ht]
\centering
\caption{Hyperparameter Search Space.}
\tiny
\label{tab:hyperparameters}
\resizebox{\columnwidth}{!}{
\begin{tabular}{ccc}
\hline
\textbf{Hyperparameter} & \textbf{Distribution} & \textbf{Search Space} \\ \hline
\ac{LR} &  Log-Uniform & $1 \times 10^{-5}$, $1 \times 10^{-3}$ \\
Batch Size & Categorical & $\{16, 32, 64\}$ \\
Optimizer & Categorical & $\{\text{Adam, SGD}\}$ \\ \hline
\end{tabular}
}
\end{table}

Among the single-model baselines (centralized approach), ResNet achieved the highest held-out performance (accuracy $0.9798$), followed closely by EfficientNet (accuracy $0.9783$). Because AIDER is class-imbalanced (e.g., the normal class is larger than others), accuracy alone may not fully capture per-class behavior. Still, it provides a stable basis for comparing centralized and federated settings. 

We report centralized and federated results primarily to contextualize accuracy and to contrast communication costs. We do not claim that federated training is inherently superior to centralized training with full data access. Small differences can arise from stochastic optimization, data ordering, and training dynamics. In our study, \ac{FL} serves as a reference paradigm that exchanges parameters across rounds, against which our method contrasts a probability-level, asynchronous alternative. 
 
Table~\ref{tab:baseline_vs_fl} compares the baseline accuracies against those obtained under an \ac{FL} setup. The results show consistent improvements across all models when moving to federated training. For example, MobileNetV3 increased from $0.9666$ to $0.9767$, and both ResNet and EfficientNet reached $0.9813$, surpassing their centralized counterparts. Even the lightweight SqueezeNet, which had the lowest baseline accuracy ($0.9465$), benefited from federated optimization, improving to $0.9557$. These results highlight that \ac{FL} not only preserves performance relative to centralized training but can also yield measurable gains, particularly in models with fewer parameters or lower baseline accuracy. 

\begin{table}[htbp]
\centering
\caption{Centralized vs \ac{FL} on \ac{AIDER}: Accuracy and Macro-F1.}
\label{tab:baseline_vs_fl}
\tiny
\resizebox{\columnwidth}{!}{%
\begin{tabular}{lcccc}
\toprule
\multirow{2}{*}{\textbf{Model}} & \multicolumn{2}{c}{\textbf{Accuracy}} & \multicolumn{2}{c}{\textbf{Macro-F1}} \\
\cmidrule(lr){2-3}\cmidrule(lr){4-5}
& \textbf{Centralized} & \textbf{\ac{FL}} & \textbf{Centralized} & \textbf{\ac{FL}} \\
\midrule
ResNet            & 0.9798 & 0.9813 & 0.9668 & 0.9681 \\
EfficientNet      & 0.9783 & 0.9813 & 0.9633 & 0.9633 \\
MobileNetV2       & 0.9659 & 0.9720 & 0.9431 & 0.9533 \\
MobileNetV3       & 0.9666 & 0.9767 & 0.9407 & 0.9584 \\
SqueezeNet        & 0.9465 & 0.9557 & 0.9087 & 0.9270 \\
\bottomrule
\end{tabular}%
}
\end{table}

\begin{figure}[!ht]
\centering
  \includegraphics[width=0.80\columnwidth]{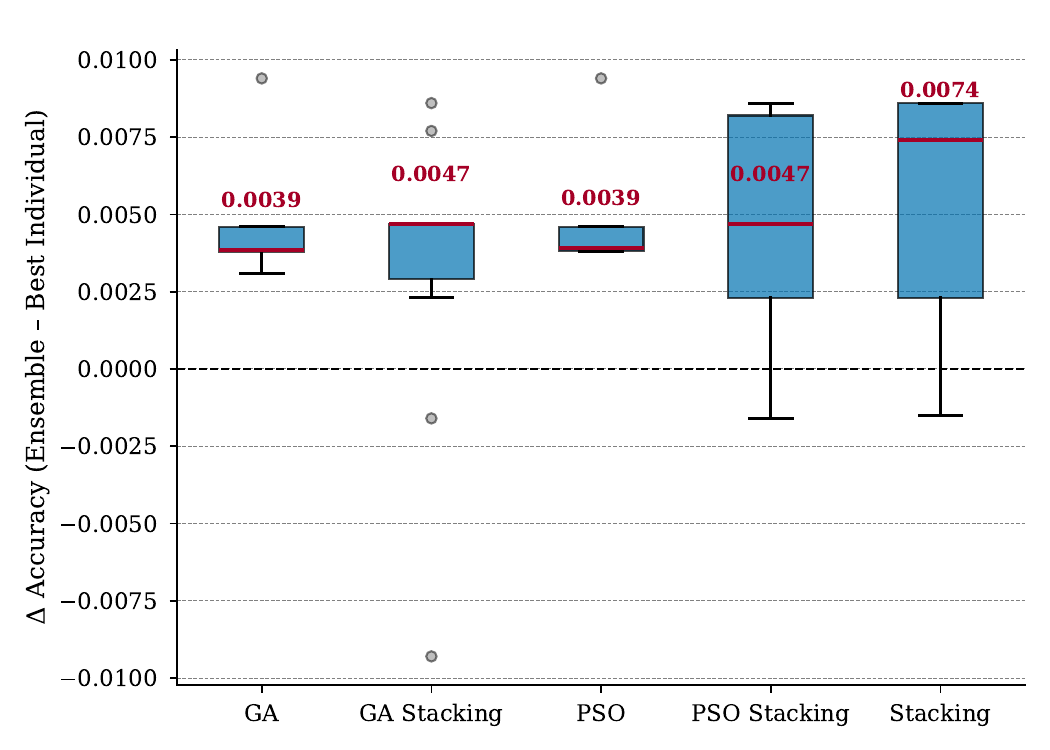}
  \caption{Distribution of accuracy gains achieved by the ensemble.}
  \label{fig:boxplot_delta}
\end{figure}

\textbf{Ensembles consistently outperform single models.} Fig.~\ref{fig:boxplot_delta} presents the distribution of the ensemble gain over the best individual model across all evaluated methods. The results clearly indicate that the proposed probability-level aggregation consistently outperforms the strongest single model in almost every case. All ensemble strategies exhibit positive median values, confirming the robustness of our approach.

While GA and PSO deliver modest but highly stable improvements, the stacking-based variants achieve larger gains at the cost of higher variability, highlighting their capacity to exploit complementary information across models. Notably, pure stacking achieves the largest median improvement of approximately 0.0074, demonstrating that ensembles built using our method can substantially improve predictive performance. Even though a few negative outliers are observed, the overall trend is decidedly upbeat, providing strong evidence that probability-level aggregation is an effective alternative to conventional \ac{FL} schemes.

\begin{figure}[!t]
\centering
  \includegraphics[width=0.9\columnwidth]{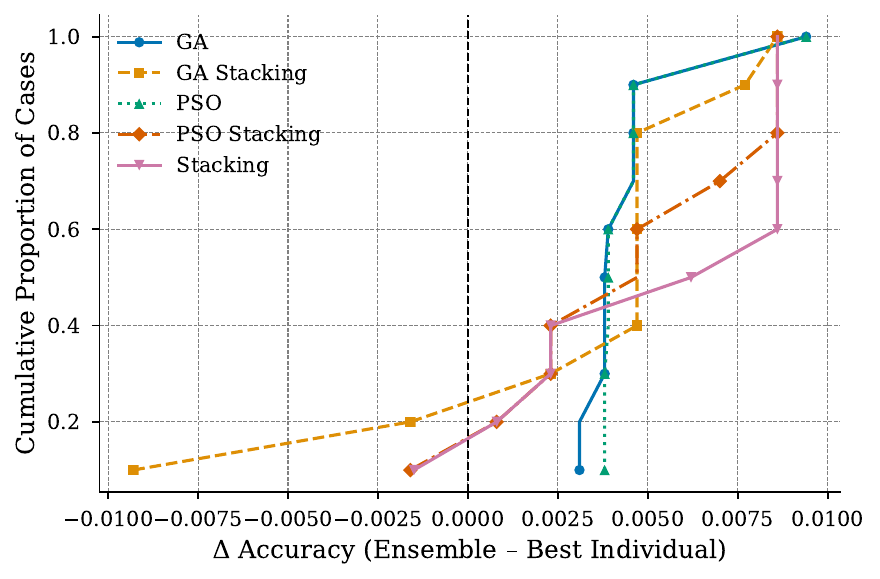}
  \caption{Cumulative distribution of ensemble gains over the best individual model.}
  \label{fig:cumulative}
\end{figure}

\textbf{Performance gains remain stable across runs.} Fig.~\ref{fig:cumulative} shows the cumulative distribution of ensemble gains over the best individual model. 
All methods lie predominantly to the right of the $x$-axis zero, confirming consistent improvements. GA and PSO yield small but stable gains, while stacking and its variants achieve larger improvements with greater variability. These results demonstrate that our method reliably outperforms the strongest centralized model.

\textbf{Model complementarity drives ensemble success.} In the ensemble stage, we first confirmed the foundational benefit of probability-level fusion. Even a simple average of a ``weak'' ensemble (MobileNetV2 + MobileNetV3 + SqueezeNet) outperformed its best individual member, achieving an accuracy of 0.9729 (compared to 0.9666), illustrating the power of variance reduction. While this trio did not surpass the top single baseline (ResNet at 0.9798), the leading ensemble (EfficientNet + MobileNetV2 + ResNet) achieved a higher score, reaching 0.9822. As shown in Table~\ref{tab:ensemble_compare}, this demonstrates that fusion is most effective when backbones are both accurate and complementary. Our approach framework builds on this principle, utilizing advanced methods such as stacking and evolutionary optimization to maximize these gains in a distributed setting, rather than relying on simple averaging.


Even modest accuracy gains can be meaningful when obtained with orders-of-magnitude lower communication, because they enable more frequent updates and broader participation under constrained links. In disaster response, this can improve timeliness and coverage without requiring synchronized rounds or parameter exchange.

\begin{table}[!ht]
\centering
\caption{Comparison of ensemble combinations.}
\label{tab:ensemble_compare}
 \renewcommand{\arraystretch}{1.3} 
\resizebox{\columnwidth}{!}{%
\begin{tabular}{@{}lccc@{}}
\toprule
\textbf{Combination}                   & \textbf{Accuracy} & \textbf{Macro F1} & \textbf{Weighted F1} \\ \midrule
EfficientNet + MobileNetV2 + ResNet    & 0.98216           & 0.96990           & 0.98204              \\
MobileNetV2 + MobileNetV3 + SqueezeNet & 0.97285           & 0.95306           & 0.97248              \\ \bottomrule
\end{tabular}
}
\end{table}

The comparative analysis highlights that ensemble models achieve highly competitive accuracy, even if they do not always surpass \ac{FL} in every case. For example, while ResNet and EfficientNet achieved $0.9813$ under federated training, the top-3 ensemble (EfficientNet + MobileNetV2 + ResNet) achieved a very similar result of $0.9822$. In contrast, weaker combinations, such as MobileNetV2 + MobileNetV3 + SqueezeNet, performed slightly worse than their federated counterparts. These results indicate that ensembles may marginally underperform in certain configurations; however, their performance remains close to that of the federated setting, particularly when strong, complementary backbones are combined.

\textbf{Our approach achieves major communication savings.} Where ensembles clearly excel is in communication efficiency. As shown in Table~\ref{tab:net_consumption}, federated training incurs tens to hundreds of megabytes of data transmission per model, whereas ensemble aggregation consistently stays below $200$~kB. For instance, ResNet required over $255$ million Bytes under \ac{FL}, compared to only $\sim 1.5\times10^5$ bytes for ensemble fusion, a difference of more than three orders of magnitude. Similar gaps are observed across all architectures. Beyond raw performance, the most striking contrast lies in network consumption. Since each client sends only a class-probability vector of size \(C\), communication grows linearly with the number of clients \(\big(O(N\!\cdot\!C)\big)\), unlike traditional \ac{FL}, which transmits the trainable parameter vector \(\theta\in\mathbb{R}^{P}\) (all weights and biases; \(|\theta|=P\)) each round \(\big(O(N\!\cdot\!P)\big)\). This lightweight design supports scalability as \(N\) and model heterogeneity increase.

\begin{table}[!ht]
\centering
\caption{Network consumption (bytes transmitted) for Ensemble vs. \ac{FL}}
\tiny
\label{tab:net_consumption}
\resizebox{0.7\columnwidth}{!}{%
\begin{tabular}{l r r}
\hline
\textbf{Model} & \textbf{Ensemble} & \textbf{Federated} \\
\hline
SqueezeNet      & 149{,}603   & 8{,}855{,}868 \\
ResNet       & 149{,}231   & 255{,}651{,}996 \\
MobileNetV3     & 149{,}231   & 18{,}421{,}932 \\
MobileNetV2     & 151{,}184   & 27{,}173{,}916 \\
EfficientNet & 148{,}649   & 48{,}672{,}804 \\
\hline
\end{tabular}%
}
\end{table}

\textbf{Practical trade-off between accuracy and cost.} These findings underscore the practical advantages of our method. In many settings, our approach matches or outperforms \ac{FL} in accuracy while reducing communication by orders of magnitude and eliminating global synchronization barriers. This makes our strategy particularly suitable for deployments with intermittent connectivity, such as Disaster Decision Support Systems (DDSS), where timely and resilient updates are as important as peak accuracy. Overall, the results indicate a favorable trade-off: although \ac{FL} achieves slightly higher peaks in some scenarios, our technique often delivers equal or superior accuracy at substantially lower communication cost. Thus, the reported numbers are directly connected to disaster response: more nodes can transmit practical knowledge more frequently at a lower per-contribution cost while maintaining decision quality.

Our approach uses an \ac{MQTT} broker as a lightweight coordination substrate in our prototype, which introduces an availability dependency in practice. This is not a conceptual requirement of our strategy; the same publish/consume pattern can be implemented with replicated brokers, fault-tolerant queues, or other messaging substrates.

\section{Concluding Remarks}\label{sec:concluding_remarks}

This study addresses the communication and synchronization bottlenecks that hinder timely disaster detection from heterogeneous edge devices. We introduced an asynchronous probability ensemble framework with feedback distillation, where clients exchange only class probability vectors and the server aggregates them via stacking or weight optimization. Experiments on AIDER show that our approach consistently improves over single backbones and attains accuracy comparable to or slightly better than federated baselines, while reducing network traffic by orders of magnitude, which supports more frequent updates and broader participation under intermittent connectivity. 

These findings suggest a practical path for DDSS deployments that must trade peak accuracy for bandwidth, latency, and resilience because decision quality is preserved without exchanging model parameters. The current design assumes a shared reference set for well-defined aggregation and distillation, and relies on a broker-based coordination substrate in our prototype. Future work will study adaptive and uncertainty-aware fusion, robustness, and fairness under non-IID data and adversarial or noisy clients, scalable operations with larger client pools, and fault-tolerant messaging to remove single-point dependencies. Overall, probability-level collaboration with feedback distillation offers a lightweight, scalable, and resource-aware alternative for distributed disaster image detection.

\section*{Acknowledgment}

The authors thank the National Council for Scientific and Technological Development (CNPq) under grant number 421944/2021-8 (call CNPq/MCTI/FNDCT 18/2021), FAPEMIG (Grant \#APQ00923-24), FAPESP MCTIC/CGI Research project 2018/23097-3 - SFI2 - Slicing Future Internet Infrastructures. FCT has also supported this work – Fundação para a Ciência e Tecnologia within the R\&D Unit Project Scope UID/00319/Centro ALGORITMI (ALGORITMI/UM). Fapes (2023-RWXSZ).

\bibliographystyle{IEEEtran}
\bibliography{references}

\end{document}